\documentclass[letterpaper, 10 pt, conference]{ieeeconf}
\usepackage{amsmath,amsfonts}
\usepackage{array}
\usepackage{enumerate}
\usepackage[utf8]{inputenc}
\usepackage[T1]{fontenc}
\usepackage{amsfonts}
\usepackage{amssymb}
\IEEEoverridecommandlockouts 
\usepackage{amsmath}  
\usepackage{breqn}
\usepackage[caption=false]{subfig}
\usepackage{textcomp}
\usepackage{stfloats}
\usepackage{url}
\usepackage{verbatim}
\usepackage{graphicx}
\usepackage{cite}
\usepackage{lipsum}
\usepackage{mathtools}
\usepackage[linesnumbered, ruled]{algorithm2e}

\usepackage{hyperref}
\hypersetup{
colorlinks=true,
linkcolor=blue,
filecolor=magenta,
urlcolor=blue,
}

\usepackage[font=footnotesize]{caption}

\title{\LARGE \bf

Hypergraph-based Coordinated Task Allocation and Socially-aware Navigation for Multi-Robot Systems
}

\author{Weizheng Wang$^{1}$, Aniket Bera$^{2}$, and Byung-Cheol Min$^{1}$ 
\thanks{$^{1}$SMART Laboratory, Department of Computer and Information Technology, Purdue University, West Lafayette, IN, USA. {\tt\small{[wang5716, minb]@purdue.edu}.}
}
\thanks{$^{2}$Department of Computer Science, Purdue University, West Lafayette, IN, USA. {\tt\small{aniketbera@purdue.edu}.}
}}

\begin{document}

\maketitle

\begin{abstract}

A team of multiple robots seamlessly and safely working in human-filled public environments requires adaptive task allocation and socially-aware navigation that account for dynamic human behavior. Current approaches struggle with highly dynamic pedestrian movement and the need for flexible task allocation. We propose Hyper-SAMARL, a hypergraph-based system for multi-robot task allocation and socially-aware navigation, leveraging multi-agent reinforcement learning (MARL). Hyper-SAMARL models the environmental dynamics between robots, humans, and points of interest (POIs) using a hypergraph, enabling adaptive task assignment and socially-compliant navigation through a hypergraph diffusion mechanism. Our framework, trained with MARL, effectively captures interactions between robots and humans, adapting tasks based on real-time changes in human activity. Experimental results demonstrate that Hyper-SAMARL outperforms baseline models in terms of social navigation, task completion efficiency, and adaptability in various simulated scenarios\footnote{The experimental videos and additional information about this work can be found at: \url{https://sites.google.com/view/hyper-samarl}.}.

\end{abstract}

\section{Introduction}

Multi-robot systems are becoming increasingly relevant in a wide range of real-world applications, including cleaning tasks in public spaces \cite{deliveryrobot,kantaros2016global, capitan2013decentralized, hu2020voronoi, jin2021distributed } like airports, monitoring operations in large facilities, and delivery robots on university campuses. In these scenarios, a team of robots must navigate shared spaces that are often filled with humans, such as pedestrians, in order to reach assigned points of interest (POIs) and complete specific tasks. As these systems become more prevalent, ensuring that robots can safely and efficiently navigate these human-filled environments while maintaining socially acceptable behavior \cite{singamaneni2024survey, eva2, eva3} presents a critical challenge.

Despite significant progress in social navigation using methods such as deep reinforcement learning (RL) \cite{wang2024multi, wang2023navistar, liu2023intention, liu2021decentralized, Xie2023DRLVO, 
Liu2024DRLSAN}, optimization algorithms \cite{opti-1, opti-2, opti-3}, and geometric theories \cite{mavrogiannis2019multi, cao2019dynamic}, integrating socially-aware navigation into multi-robot systems remains a difficult problem. The challenge is primarily due to the need for robots to navigate in highly dynamic environments filled with dense human activity, where they must adhere to social norms and avoid collisions. Whether the robots are part of a team performing cleaning tasks in a busy airport or monitoring large facilities, optimizing socially-compliant path planning is crucial to improving both the safety and performance of these systems.


\begin{figure}[!t]
\centering
\includegraphics[width=1\columnwidth]{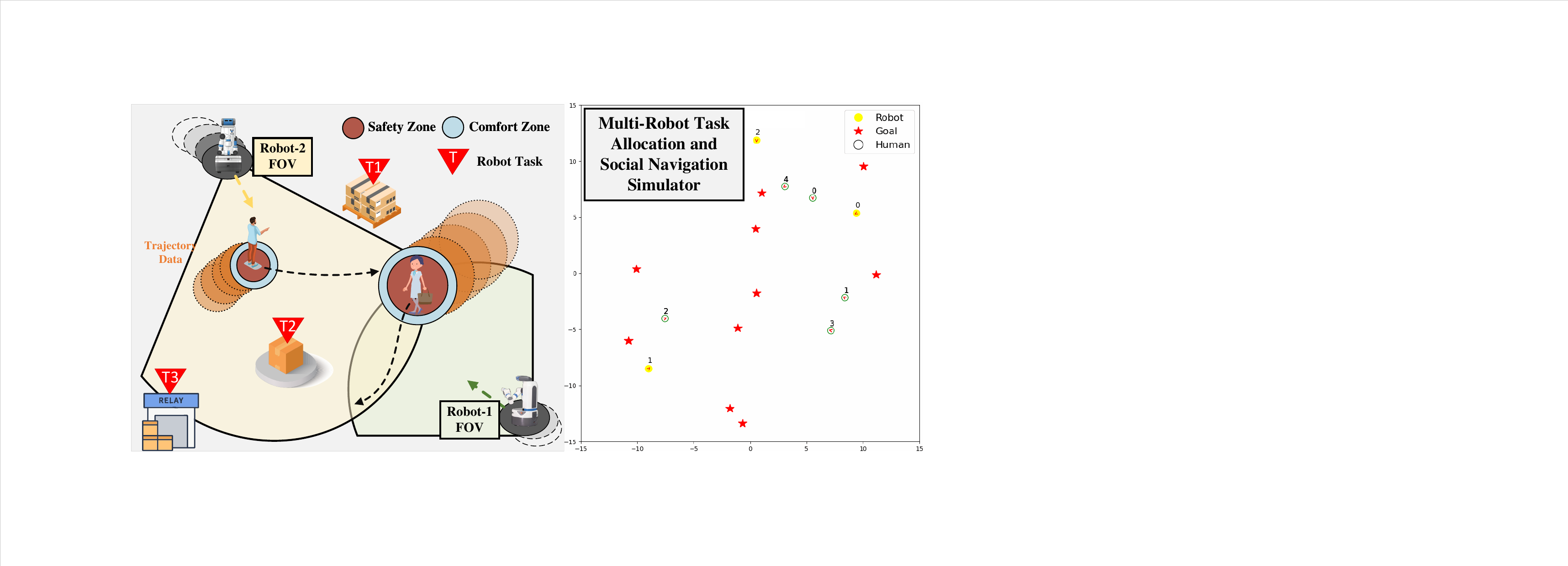}
\vspace{-2pt}
\caption{An illustration of  multi-robot task allocation and social navigation task and simulator: mobile robots are engaging in cooperative navigation toward POIs in a human-filled environment while adhering to social norms.}
\vspace{-20pt}
\label{fig:F1}
\end{figure}

This study addresses the problem of multi-robot navigation in complex, human-populated environments, as illustrated in Fig.~\ref{fig:F1}. The core challenge lies in autonomously assigning POIs to multiple robots while determining feasible and socially-aware paths that allow the robots to navigate around humans. Although task allocation for multi-robot systems has been well explored \cite{farinelli2017distributed, lei2022multitask, elfakharany2020towards, mataric2003multi, jose2016task, alitappeh2022multi, chakraa2023optimization}, many existing approaches assume static obstacle configurations \cite{obstacle2, obstacle1, obstacle3} and struggle to adapt to the dynamic and unpredictable nature of human environments. Human movement frequently alters the landscape in real time, making static strategies insufficient for allowing robots to effectively reach their destinations. To overcome these limitations, we extend socially-aware navigation algorithms \cite{wang2024multi} to improve human-robot cooperation, with a specific focus on enabling multi-robot systems to operate effectively in dynamic, human-filled public spaces.

As human environments are inherently dynamic, new challenges arise in the form of unpredictability. Fixed task allocation strategies are often inadequate in these contexts, requiring robots to adopt more flexible and adaptive behaviors. For instance, if a pedestrian walks near a robot’s assigned destination, the robot must be capable of dynamically adjusting its task assignment in real time to avoid potential collisions or disruptions. To address these challenges, we propose an advanced hypergraph-based neural network that dynamically adjusts task allocation based on real-time environmental changes. This enables multi-robot systems to adapt quickly to changes in human behavior, ensuring that they navigate human-populated environments efficiently and in a socially compliant manner.

The main contributions of this paper are as follows: 
\begin{itemize}
   \item We propose Hyper-SAMARL, a novel \textbf{Hyper}graph-based multi-robot task allocation and \textbf{S}ocially-\textbf{A}ware navigation framework using \textbf{M}ulti-\textbf{A}gent \textbf{R}einforcement \textbf{L}earning. In this framework, POIs are dynamically assigned to robots, including the specific order of visitation, while ensuring non-collision paths that adhere to social norms in human-populated environments.

    \item Hyper-SAMARL features a hypergraph diffusion mechanism, which converges high-order correlations between robots, humans, and POIs, enabling adaptive task allocation and dynamic navigation behaviors in response to changes in the environment.

    \item Extensive experiments are conducted to evaluate the framework’s performance in various dynamic, human-populated environments. These experiments demonstrate the effectiveness of Hyper-SAMARL in ensuring socially-compliant navigation, efficient task completion, and adaptability to complex conditions.


  
\vspace{-2mm}
\end{itemize}

\section{Related Works}

Socially-aware robot navigation is a foundational topic in robotics and has inspired numerous studies \cite{luber2012socially, singamaneni2024survey, moller2021survey, charalampous2017recent}. Current approaches to social navigation can generally be classified into two categories: those that couple HRI inference with path planning and those that decouple them. Early efforts treated pedestrians as static obstacles \cite{montemerlo2003perspectives, fox1998hybrid}, leading to unnatural and uncomfortable robot behaviors in human-shared environments. While decoupled planners have been successfully deployed in some scenarios \cite{cao2019dynamic, du2011robot, kim2016socially}, their lack of cooperative collision avoidance and planning adaptability often results in the "freezing robot problem" \cite{freezing} in more complex and crowded environments. 

More recent work focuses on learning-based approaches that couple HRI inference with path planning \cite{wang2023navistar, wang2024multi, SunM-RSS-21}, leveraging neural networks or preference distribution to approximate human preferences and cooperation tendencies. For multi-agent scenarios, studies such as \cite{han2020cooperative, long2018towards} introduced training paradigms that integrate multi-robot systems into a single policy network using RL algorithms like PPO. More recently, \cite{escudie2024attention, wang2024multi} extended single-agent RL to multi-agent reinforcement learning (MARL) with the MARL benchmark multi-agent proximal policy optimization (MAPPO) algorithm to train graph-based or transformer-based neural networks for multi-robot social navigation. Despite these promising developments in implicit coordination for multi-robot social navigation, the integration of advanced task allocation and navigation strategies remains underexplored. To address this gap, we frame the multi-robot socially-aware navigation (MR-TASN) task as a decentralized partially observable semi-Markov decision process (Dec-POSMDP) problem \cite{POSMDP} to couple task allocation with social navigation. 

Path planning and task allocation in multi-robot systems are inherently complex, requiring coordination and self-learning in dynamic and uncertain environments. Several approaches have been proposed to tackle these challenges. For instance, \cite{martin2023multi} introduced a cluster-based task allocation approach method grounded in game theory. \cite{mahato2023consensus} formulated the multi-robot task allocation problem as an optimization task using a graph structure. Also, \cite{motes2023hypergraph} constructed a hypergraph to represent the correlations between robots and objects in the environment. However, many of these approaches assume static obstacle configurations \cite{obstacle2, obstacle1, obstacle3} or fail to address the adaptability of task re-assignment in dynamic, human-filled environments. 

In this work, we explore the feasibility of multi-robot task allocation in dynamic, human-populated environments, framing it as an MR-TASN problem. We leverage hypergraph diffusion to dynamically adjust both task allocations and navigation behaviors in response to real-time environmental changes. Our proposed framework, Hyper-SAMARL, provides a more adaptive and robust solution to the challenges of task allocation and socially-aware navigation in multi-robot systems.

\section{Preliminary}
\label{sec:Preliminary}

\subsection{Hypergraph Dynamic Relational Reasoning}

We model the environmental dynamics of MR-TASN tasks using a hypergraph construction $\mathcal{G} = (\mathcal{V,E})$, where $\mathcal{V} = \{{v_{ 1}},{v_{ 2}},\cdots,{v_{ N}}\}$ is the vertex set representing $N$ objects, including robots, pedestrians, and POIs. The relationships among these objects are captured by hyperedge set $\mathcal{E} = \{{e_{ 1}},{e_{ 2}},\cdots,{e_{M}}\}$. The hyperedge weight matrix $\mathbf{W} = diag({{w}_{e_1}},{{w}_{ e_2}},\cdots,{{w}_{ e_M}}) \in \mathbb{R}^{M \times M}$, along with the hypergraph vertex features $\{X_1, \cdots, X_N\}$, are estimated based on the spatial-temporal dependencies of objects. 

The vertex degree matrix $\mathbf{D}_V \in \mathbb{R}^{N \times N}$ is diagonally composed of the vertex degrees $\delta_v=\sum_{e\in \mathcal{E}} w_{(e)}\mathbf H(v,e)$. Similarly, the hyperedge degrees $\varsigma_e=\sum_{v\in \mathcal{V}}\mathbf H(v,e)$ fill the hyperedge degree matrix $\mathbf{D}_E \in \mathbb{R}^{M \times M}$, where the hypergraph incident matrix $\mathbf{H} \in \mathbb{R}^{N \times M}$ is defined as follows: $\mathbf{H}(v,e) = 1, ~\text{if}~ v \in e;~ \mathbf{H}(v,e) = 0, ~ \text{otherwise}.$ 

We initialize the hypergraph of MR-TASN based on spatial-temporal transformer features $\mathbf{X}_{ST}$ and Euclidean distance attributes. Subsequently, the hypergraph diffusion framework is employed to update the fixed correlations, leveraging both vertex attributes and the edge structure of its clique-expanded graph. Eventually, the hypergraph features are used for task allocation and navigation execution, which are trained using MARL.

\subsection{Markov Decision Process Formulation}

Drawing inspiration from \cite{wang2024multi,POSMDP}, the MR-TASN task is formally constructed as a Dec-POSMDP with the tuple $\langle \mathcal S, \mathcal U,\mathcal A,\mathcal{O},\mathcal{P} ,\mathcal R, \mathbf{R}, \mathcal{C}, \mathcal S_0, \gamma \rangle$. The joint state is $\hat{s}_t = [s_t^{r_1}, \cdots, s_t^{r_n}] \in \mathcal{S}$, and the joint observation is $\hat{o}_t = [o_t^{r_1}, \cdots, o_t^{r_n}] \in \mathcal{O}$. Hyper-SAMARL converts local observations into macro-action (MA) $\hat{u} = [p_{gx}^{(1,\cdots,n)}, p_{gy}^{(1,\cdots,n)}] \in \mathcal{U}$ via robot policy $\hat\pi$ to assign POI positions $\{p_{gx}^{}, p_{gy}\}$ and guide the generation of velocity-based local-action (LA) $\hat{a} = [v_x^{(1,\cdots,n)}, v_y^{(1,\cdots,n)}] \in \mathcal{A}$. Specifically, the robots' MA are updated at each decision-making timestep $t_k \in \{t_0, \cdots, t_K\}$, while the robots' LA are sequentially executed during the intervals between decision-making timesteps $t \in  [t_{k}, t_{k+1})$. 

The LA reward function for a total of $n$ robots is given by $\hat{\mathbf{R}}:\mathcal{S}^{}\times\mathcal{U}^{ }\mapsto\mathbb{R}^{ n}$, which is obtained by each timestep. Moreover, the MA reward function is defined as the expected maximization of the sum of the LA rewards: $\hat{\mathcal{R}}({\hat{s},\hat{u}}) = \mathbb{E} [\textstyle\sum_{ t=0}^{ T} \gamma^{ t} \hat{\mathbf{R}}(\hat{ s}_{ t},\hat{ a}_{ t})|\hat{ a}_{ t} \sim \hat{ u}( H_t) ]$ over the period $T = t_{k+1} - t_{k}$. Additionally, various environmental dynamics, such as collision occurrences and arrival at POIs, are encoded into the system's conditional function $\mathcal{C}$, which is utilized to detect key events in the environment. The $\gamma \in [0, 1]$ is the discounted factor. For more details, please refer to \cite{wang2024multi, POSMDP}.

Formally, both observed and unobserved states are involved in the individual state of a pedestrian and robot, denoted as $ \mathbf{s}_{t} = [ \mathbf{s}^{o}_{{t}}, \mathbf{s}^{uo}_{{t}}]$. The observed state, $ { \mathbf{s}^{o}_{{t}}} = [ p_{ x}, p_{ y}, v_{{x}}, v_{{y}},\rho]$, contains public information such as current location, velocity, and radius. On the other hand, the unobserved state $\mathbf{s}^{{uo}}_{{t}}=[ {g}_{{x}},{g}_{{y}},v_{{pref}},{\theta}]$, includes private information such as target destination and policy strategy. The local observation is defined as $\hat{o}_t = [{s}^{r_1}, \hat{s}^{o_1}, \cdots, {s}^{r_n}, \hat{s}^{o_n}]$.

Robots are initialized according to the state distribution $\mathcal{S}_{0}$ and the original MA $u_{t_0}$ at the beginning of each episode. The robots then update the sequence of POI positions using MA at each decision-making timestep ${t_k}$, guiding the generation of LA $\{ a_t,\cdots, a_{t+T} \}$. Meanwhile, the environment calculates the reward feedback for the state transition and determines the next state based on the transition probability $\mathcal{P}$. This process either terminates or completes based on the outcome of the conditional function $\mathcal{C}$.

\begin{figure*}[!t]
\centering
\vspace{-5pt}
\includegraphics[width=1\linewidth]{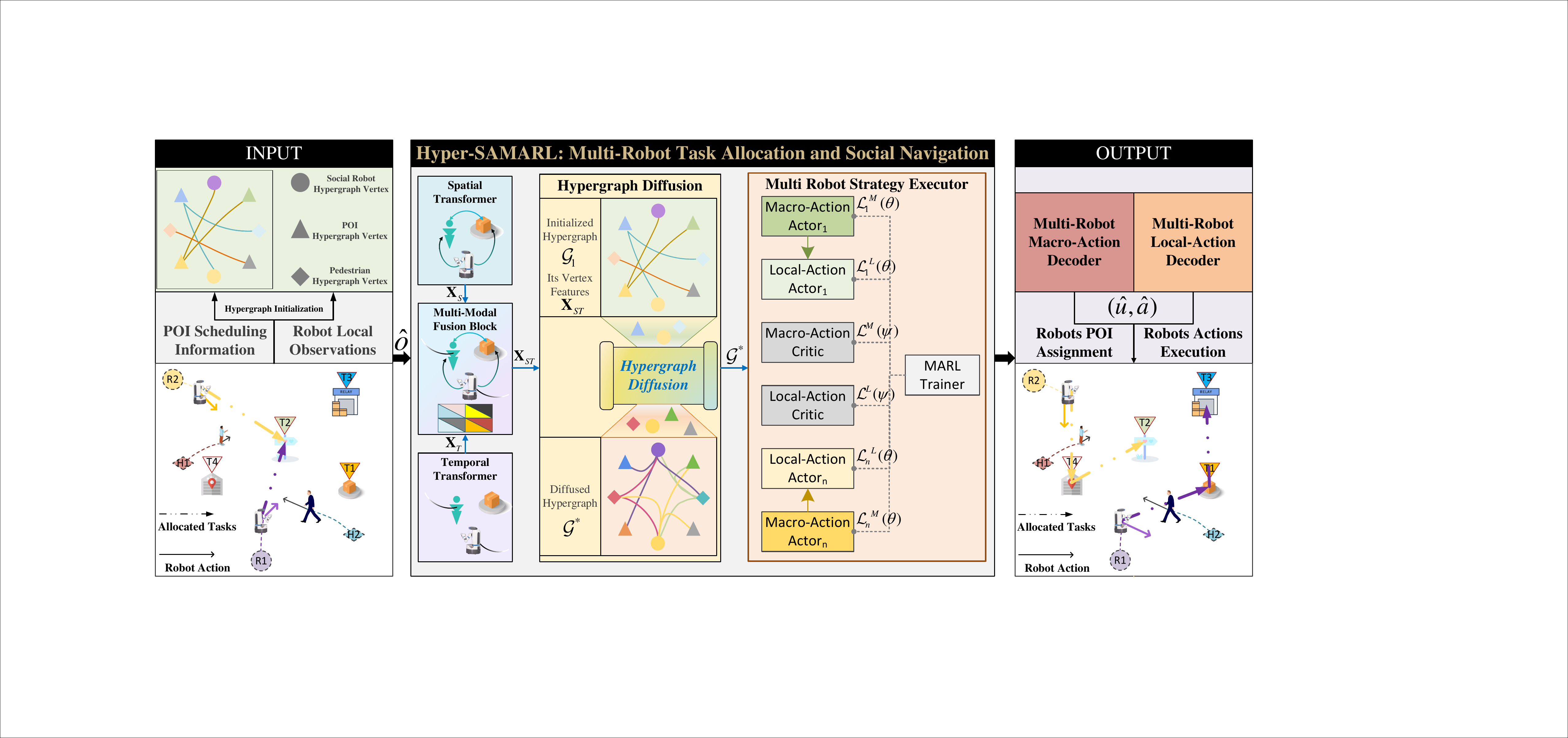}
\vspace{-5pt}
\caption{Hyper-SAMARL Architecture: First, task information related to points of interest (POIs) and robot observations are encoded by the positional embedding encoder of the transformer as spatial-temporal input. Next, the hypergraph for MR-TASN is initialized using attention-based vertex features and Euclidean-based hyperedge features. The hypergraph diffusion mechanism is then employed to propagate vertex features across the hypergraph structure, ensuring balanced hypergraph dynamics. Finally, the diffused hypergraph features are decoded by the robot policy to generate macro-actions (MA) and local actions (LA), which are trained using the MAPPO algorithm.}
\vspace{-15pt}
\label{fig:F2}
\end{figure*}

\subsection{Robots Task Allocation $\&$ Social Navigation Statement}

The objective $\mathcal{J}$ of MR-TASN, where $n$ robots explore a set of $p$ POIs alongside $m$ pedestrians in an open space, can be defined as follows:
\vspace{-4pt}
\begin{equation}
\begin{aligned}
&\mathcal{J} = \mathop{\arg\min}\limits_{\forall \{\tau \in \mathcal{T}, {i} \in  n\}}\; \textstyle\sum_{{i}=1}^{{n}} c_{ i}^{ t}(\tau_{ i}) + c_{ i}^{ p}(\tau_{ i}) + c_{ i}^{ s}(\tau_{ i})
\\
&=\mathop{\arg\max}\limits_{\hat{\pi}}\mathbb{E}[\textstyle\sum_{ k=0}^{ K} \gamma^{{t_k-t_0}} {\hat{\mathcal{R}}}({\hat{s}_{t_k}},{\hat{u}_{t_k})}| (\hat{\pi},{\hat{s}_0})]\\
&\text{s.t.} \;\; \forall {i,j} \in [r_{ 1},r_{ n}],\forall {h} \in [h_{ 1},h_{ m}], \forall {g} \in [POI_{ 1},POI_{p}]\\
&d_{ i,j} > \rho_{ i} + \rho_{ j} , \, i \neq j; \;\;d_{ i,h} > \rho_{ i} + \rho_{ h}; \;\; d_{ i,g} < \rho_{ i};\; \hat{u} = \emptyset 
\end{aligned}
\end{equation}
\noindent where $\tau \in \mathcal{T}$ denotes a feasible path instance from a clear path region that aims to minimize the sum of cost functions $c^{\rm t}(\cdot), c^{p}(\cdot) \text{and } c^{\rm s}(\cdot)$, representing time efficiency cost, POI exploration cost, and socially-compliance cost, respectively. These costs are then relaxed to maximize the expected macro-action reward $\mathcal{\hat{R}}$ subject to the objective constraints, which include distance constraints between robots $d_{i,j}$, between robots and humans $d_{i,h}$, and between robots and POIs $d_{i,g}$, as well as the condition that no POI exploration tasks remain, represented by $\hat{u} = \emptyset$.


\section{Methodology}

\subsection{Spatial-Temporal Transformer}

The representation of spatial-temporal social context has been extensively studied in various work \cite{wang2023navistar, wang2024multi, wang2024hyper} to address human-robot interactions or human-human interactions. Inspired by \cite{wang2024hyper}, we approximate hypergraph vertex features $\mathbf{X}_{ST}$ by capturing the spatial-temporal environmental dependencies between multi-robots, pedestrians, and task destinations, leveraging transformer-based spatial-temporal encoders. The vanilla multi-head attention mechanism \cite{vaswani2017attention} is defined as follows:
\begin{equation}
\vspace{-1pt}
\begin{aligned}
        &\operatorname{Atten}(\mathbf{O})= \operatorname{Atten}\left( \mathbf{{Q}}, \mathbf{{K}}, \mathbf{{V}}\right)={\mathrm{softmax} (\frac{ \mathbf{{Q}} \mathbf ({{\mathbf{K}} })^{\top}}{\sqrt{ {d}_{{h}}}})  \mathbf{{V}} }  
        \\
        &\operatorname{Multi}(\mathbf{O})=\operatorname{Multi}\left(\mathbf{{Q}} ,\mathbf{{K}} ,\mathbf{{V}} \right)=f_{{fc}}( {head}_1,\cdots, {head}_{{h}} ) ; \\
         &{head}_{(\cdot)}=\operatorname{Atten}{(\cdot)}\\
        &\{\mathbf{{X}}_{{S}}; \mathbf{{X}}_{{T}}\}=  \operatorname{Trans}_{{}}( \operatorname{Multi_S}(\cdot);  \operatorname{Multi_T}(\cdot) | \{\mathbf{O}_{R}, \mathbf{O}_{T}\})\\
\end{aligned}
\end{equation}
\noindent where $\mathbf{Q}, \mathbf{K}, \mathbf{V}$ are the transformer's query, key, and value matrices, respectively, and $d_h, h$ represent the transformer dimension and the number of attention heads.

In detail, agents' spatial coordination interactions and individual temporal dynamics dependencies are captured by the spatial encoder and temporal encoder, respectively. Subsequently, the heterogeneous spatial-temporal features are aligned using the multi-modal transformer \cite{tsai2019multimodal} to capture the fused spatial-temporal features of the objectives $\mathbf{X}_{ST}$ as follows:
\begin{equation}
\begin{aligned}
&\operatorname{CMAtten}(\mathbf{X}_{U}) = \operatorname{Multi}(\mathbf{{Q}}^{{}}_{{U}}, \mathbf{{K}}^{{}}_{{U'}}, \mathbf{{V}}^{{}}_{{U'}})\\
&\mathbf{{X}}_{{ST}}^{} =  \operatorname{{Trans}_{Mul}} 
(\operatorname{CMAtten}(\{\mathbf{{X}}_{{S}}, \mathbf{{X}}_{{T}} \}))
\end{aligned}
\end{equation}
\noindent where $U \in \{S, T\}$ represents the unit modality, and $U' \in \{S, T\} - U$ denotes the complementary set.

\subsection{Hypergraph Diffusion}

The hypergraph of MR-TASN is initialized to explicitly represent the correlation between multiple objectives, based on spatial Euclidean distance similarity \cite{wang2024hyper}. Formally, the MR-TASN hypergraph is initialized by grouping spatially closest objectives (robots, pedestrians, and POIs), which are connected as hyperedges. In the initialized hypergraph, the environmental dynamics of the social context are encoded into hypergraph vertex features, while the spatial distance information between objectives is stored in the hyperedges. Once the hypergraph of MR-TASN is constructed, we embed the vertex features based on the spatial-temporal dependencies $\mathbf{X}_{ST}$. Due to the lack of exhaustive HRI relational reasoning in the hypergraph initialization, Hyper-SAMARL employs a hypergraph diffusion network to propagate and reorganize the vertex features and hypergraph structure, facilitating effective task allocation and smooth, socially acceptable behavior execution, as illustrated in Fig.~\ref{fig:F2}.

Typical hypergraph diffusion \cite{prokopchik2022nonlinear, fountoulakis2021local, wang2023equivariant} restructures singular and unbalanced vertex features to achieve convergence, ensuring balanced hypergraph dynamics. This process follows the paradigm of molecules spreading from areas of high concentration to equilibrium. In this context, we formulate the optimization target for hypergraph diffusion in MR-TASN tasks, considering the interactive attributes of both vertex and hyperedges,  as follows:
\begin{equation}
\begin{aligned}
& \operatorname*{arg\,min}_{\mathcal{G}} {\sum}_{v \in \mathcal{V}} f_V(\mathbf{X}_{ST}; \mathcal{G}) + {\sum}_{e \in \mathcal{E}} f_E(\mathbf{X}_{ST}; \mathcal{G})   \\
\Rightarrow & \operatorname*{arg\,min}_{\mathcal{G}} \Big{\lVert} \mathcal{G} - \frac{\mathbf{E}_{}}{\varphi{(\mathbf{E}_{})}} \Big{\rVert} ^{2} + \frac{\alpha}{1-\alpha} \cdot \Omega(\mathcal{G})
\end{aligned}
\end{equation}

\noindent where $f_V(\cdot), f_E(\cdot)$ denote the hypergraph vertex and hyperedge potential functions, respectively, $\alpha \in [0,1]$ represents the regularization coefficient, $\mathbf{E} = \mathbf{X}_{ST} / \varphi(\mathbf{X}_{ST}) $ represents the shifted and scaled hypergraph vertex features, and $\varphi$ denotes the normalized hyperedge aggregation feature.
\begin{equation}
\varphi( \mathcal{G} ) = 2\cdot\sqrt{\sum_{e \in \mathcal{E}} w(e) \Big{\Vert} \mu( \{ \frac{\mathcal{G}_j}{\sqrt{\delta_{j}}} \} ) \Big{\Vert}^2}
\end{equation}
\noindent where $\mu$ denotes the mean of the hyperedge features.
\begin{equation}
\mu( \{ \frac{\mathcal{G}_j}{\sqrt{\delta_{j}}} \} ) = \sigma (\mathbf{H}^{\top} \cdot \varrho(\mathbf{D}_{V}^{-\frac{1}{2}} \mathcal{G}))
\end{equation}
\noindent where $\sigma$ and $\varrho$ represent the $p$-based nonlinear operators.

Precisely, we introduce the hypergraph regularization term \cite{prokopchik2022nonlinear} based on the $L^2$ regularization \cite{zhou2006learning, gao2022hgnn} to minimize the generalized variance of vertex embeddings, rather than the distance between vertex pairs, on the hypergraph $\mathcal{G}$ as follows:
\begin{equation}
\begin{aligned}
\Omega_{L^2}(\mathcal{G}) & = \sum_{e \in \mathcal{E}} \sum_{i, j \in {e}} \frac{w(e)}{d(e)} \Big{\lVert} \frac{\mathcal{G}_i}{\sqrt{\delta_{i}}} - \frac{\mathcal{G}_j}{\sqrt{\delta_{j}}} \Big{\rVert}^{2}\\
\Rightarrow \Omega_{hg}(\mathcal{G}) & = \sum_{e \in \mathcal{E}} \sum_{i,j \in e} w(e) \Big{\lVert} \frac{\mathcal{G}_i}{\sqrt{\delta_{i}}} - \mu_{}( \{ \frac{\mathcal{G}_j}{\sqrt{\delta_{j}}} \} )
\Big{\rVert}^{2}.
\end{aligned}
\end{equation}

Hyper-SAMARL employs a non-linear hypergraph diffusion framework to address the optimization problem of task allocation and social navigation. The diffusion process, iterated over $k = [1, \cdots, \mathcal{K}]$ re-evaluates hypergraph features by aggregating vertex interactions, following the structure of hyperedges in their clique-expanded graph. Ultimately, Hyper-SAMARL reconstructs the hypergraph to ensure balanced hypergraph dynamics, approximating the optimal  vertex embeddings for adaptive multi-robot cooperative task allocation and HRI-informed social navigation, as follows:
\begin{equation}
\begin{aligned}
    & q(\mathcal{G}^{(\mathcal{K})} : \mathcal{G}^{(1)} |~\mathcal{N}) = \textstyle{\prod_{k=1}^{\mathcal{K}}} q(\mathcal{G}^{(k+1)} ~|~ \mathcal{G}^{(k)}) \\
    & q(\mathcal{G}^{(k+1)}|~ \mathcal{G}^{(k)} ; \mathcal{N}) = \frac{\alpha \mathcal{N}(\mathcal{G}^{(k)}) + (1-\alpha)\mathbf{E}}{\varphi(\alpha \mathcal{N}(\mathcal{G}^{(k)}) + (1-\alpha)\mathbf{E})}
\label{Diffusion}
\end{aligned}
\end{equation}
where the nonlinear diffusion map $\mathcal{N}$ is defined using the Laplacian operator, as follows:
\begin{equation}
\mathcal{N}(\mathcal{G}) = \eta ( \mathbf{H} \mathbf{W} \cdot \sigma (\mathbf{H}^{\top} \cdot \varrho(\mathbf{D}_V^{-\frac{1}{2}} \mathcal{G}))).
\end{equation}

To better account for the non-stationary nature and uncertainty in the multi-agent scenario, which arise from the influence of other agents' actions, the hypergraph diffusion's nonlinear operators $\sigma, \varrho,$ and $\eta$ are parameterized using a multilayer perceptron (MLP). This enhances the learnability of nonlinear features during the diffusion process, both from vertex-to-edge and edge-to-vertex, as follows:
\begin{equation}
\begin{aligned}
& \varrho(X_I) = ({X_I})^p ~;~\sigma(X_I) = MLP((\mathbf{D}_E^{-1}X_I)^{1/p})\\
& \eta(X_I) = MLP(\mathbf{D}_V^{-\frac{1}{2}} X_I).
\end{aligned}
\end{equation}

Eventually, the hypergraph diffusion process converges according to the threshold function $f_{th}((\Vert \mathcal{G}^{(k + 1)} - \mathcal{G}^{(k)} \Vert ~/~ \Vert \mathcal{G}^{(k + 1)} \Vert) \leq \varepsilon)$, where $\varepsilon$ represents the approximated tolerance factor. The stationary point of the diffusion process is sensitive to the configuration of the RL reward function, ensuring balanced and optimized resource assignments in the system. This is achieved by considering the interactions across the attributes of vertices, hyperedges, and hypergraph constructions. 

Hyper-SAMARL decodes the final diffused feature $\mathcal{G}^{(*)}$ through the robot MA actor and LA actor to perform multi-robot task assignment and cooperative action generation via the Dec-POSMDP framework. The convergence of the nonlinear hypergraph diffusion process is supported by the findings in\cite{prokopchik2022nonlinear}.

\subsection{Multi-Agent Reinforcement Learning}

In this work, we address the MR-TASN task as a Dec-POSMDP, utilizing the MAPPO \cite{yu2022surprising}, which serves as a benchmark for MARL. Our approach builds on the state-of-the-art socially-aware multi-agent reinforcement learning (SAMARL) \cite{wang2024multi}, a leading multi-robot social navigation planner. As shown in Algorithm~\ref{alrorithm1}, the task coordination and cooperative navigation behaviors are trained using the centralized training and decentralized execution (CTDE) paradigm. This approach leverages global information during training to mitigate the non-stationary nature of multi-agent scenarios.

\begin{algorithm}[h]
 Initialize parameters $(\theta, {\theta'}, \psi, {\psi'}, p_{star}^{}, p_{diff}^{} )$\;
\While {$step\leq step_{max}$}{
	  Initialize data buffer $\mathcal{D}=\left\{\right\}$\;
      \For{$i=1~ to ~batch\_size$}{
           Reset the environment\; 
           Create $N$ empty caches $C = [[\ ],\dots,[\ ]]$\;
      		\For{$t_k, (k=0~ to ~K)$}{
      			\For{$all ~agents ~i=1 ~to ~N$}{
      			{\textbf{if} $\text{agent} ~i ~\text{updates} ~\text{MA}\; {{u}_{t_k}^{i}}$} on $t_k$:
                    {
                    $\mathbf{X}_{ST}^{i} = \mathrm{Trans_{STAR}}(\mathbf{O}_{R}^{i}, \mathbf{O}_{T})$\;

                    Initialize the Hypergraph $\mathcal{G}^{i, (0)}_{t_k}$\;
                    
                    \While{$f_{th} = 1$}
                    {
                    $\mathcal{G}_{t_k}^{i,*} = q(\mathcal{G}^{(\mathcal{T})} : \mathcal{G}^{(1)} |~\mathcal{N})$
                    }
                    
                    $\vartheta_{t_k} = \mathbf{V}_{\psi}(\hat{s}_{t_k}; {\psi}, \hat{p}_{star}, \hat{p}_{diff}, \hat{H}_{t_k})$\;
      		    $C^{i}+=[s_{t_k-1}^{i},o_{t_k-1}^{i}, {{u}_{t_k-1}^{i}},p_{star}^{i},$ $ {{p}_{diff}^{i}, H_{t_k}^{i},{\mathcal{R}}_{t_k}^{i}}, s_{t_k}^{i},o_{t_k}^{i}]$\;
      				  Update macro action ${{u}_{t_k}^{i}} \sim \pi_{\theta}^{i}( \mathcal{G}_{t_k}^{i,*}$)\;
      			}
      		}   
        Execute $\mathrm{a}_{t}^{i}\sim {\pi}_{\theta'}^{i} (o_{t_k}^{i},{{u}_{t_k}^{i}};\theta', \mathcal{G}_{t_k}^{i,*}, H_{t_k}^{i})$\;
        $\vartheta_{t} = \mathbf{V}_{\psi'}(\hat{s}_{t},\hat{u}_{t_k}; \psi', \hat{p}_{star}, \hat{p}_{diff},\hat{H}_{t})$\;        
      	}
      Compute reward and insert data into $\mathcal{D}$\;
      }
	 	Update $(\theta, \theta', \psi ,\psi', p_{star}^{}, {p}_{diff})$ with MAPPO\;
}      
\caption{Hyper-SAMARL Training Procedure}
\label{alrorithm1}
\end{algorithm}
\setlength{\textfloatsep}{0.15cm}

Apart from that, Hyper-SAMARL utilizes a macro actor-critic and a local actor-critic network to formulate the Dec-POSMDP. The macro-actions of agents $ \hat{u}_{t_k} = [\hat{g}_{t_k}^{1}, \cdots, \hat{g}_{t_k}^{Q}]$ are updated at each decision-making timestep $t_k \in [t_0, \cdots, t_K]$ as a sequence of target destinations. Meanwhile, local-actions, $\hat{a}_t = [\hat{v}_{t}^{}]$, are generated to directly control the robot using velocity vectors for collision avoidance during the period $t \in [t_k, t_{k+1}]$. 

In detail, robot local observations are first encoded by a spatial-temporal transformer to capture HRI context as hypergraph vertex features. The hypergraph is initially constructed based on spatial Euclidean distance similarity. The comprehensive correlations between objectives are then inferred via hypergraph diffusion for macro-action of task allocation, while local-actions are guided by both the system feature $\mathcal{G}^{*}$ and spatial-temporal dependencies $\mathbf{X}_{ST}$. Additionally, both the transformer $p_{star}$ and hypergraph neural networks $p_{diff}$ are backpropagated through the multi-robot actor-critic network using the following loss functions:

\setlength{\abovedisplayskip}{-1pt}
\begin{equation}
\begin{aligned}
&\mathcal{L}(\theta) = {\sum\limits}_{\rm i=1}^{\rm N} \mathbb{E}_{{o}\sim \mathcal{O}, {a}\sim \mathcal{A}}[\min(\frac{\pi_{\theta}({a}^{{i}}|{o}^{{i}})}{\pi_{\theta_{old}}({a}^{{i}}|{o}^{{i}})} \mathbf{\hat{A}}^{\rm i},\\ 
& ~~~~~~~~~ clip(\frac{\pi_{\theta}({a}^{{i}}|{o}^{{i}})}{\pi_{\theta_{old}}({a}^{{i}}|{o}^{{i}})} ,1\pm\epsilon)
\mathbf{\hat{A}}^{ i}) + \kappa\mathbf{\hat{P}}^{ i}]
\\
&\mathcal{L}(\psi) = {\sum\limits}_{ i=1}^{ N} \mathbb{E}_{{s}\sim \mathcal{S}} [\max((\mathbf{V}_{\psi}({s}^{ i})-\mathbf{R}^{{i}})^{2}, \\
& ~~~~~~~~~ ({clip}((\mathbf{V}_{\psi}({s}^{ i}),\mathbf{V}_{\psi_{old}}({s}^{ i}) \pm \epsilon')-\mathbf{R}^{ i})^{2}]
\end{aligned}
\end{equation}
\setlength{\abovedisplayskip}{5pt}

\noindent where $\mathbf{\hat{A}}$ is the advantage function, computed using Generalized Advantage Estimation (GAE) \cite{schulman2015high}, and $\mathbf{\hat{P}}$ represents the policy entropy, with an entropy coefficient hyperparameter $\kappa$.



The joint reward function $\mathbf{\hat{R}}$ is calculated from each individual reward function $\mathbf{R}({s}_{ t}^{ i}, {a}_{ t}^{ i})$, which is defined as follows:
\begin{equation}
	\mathbf{R}({s}_{ t}^{ i}, {a}_{ t}^{ i})=
	\begin{cases}
	100,& \text{if } \forall r\in[1,\mathrm N] \ dis_{\rm r,g}<\rho_{\rm r}\\    
     25,& \text{else } dis_{\rm i,g}<\rho_{\rm i}\\
	    -100,& \text{else } \rm{s_t} \in \mathcal{C}_{collision}(\mathbf{s}_{\rm t}^{\rm i},\mathbf{a}_{\rm t}^{\rm i}) \\
            -100, & \text{else } t = t_{max}, \exists ~ u^{i} \neq \emptyset\\
	    max(\frac{-1}{dis_{i,h}},-5), & \text{else } dis_{\rm i,h} \leq 0.45 $\cite{rios2015proxemics}$ \\
        \frac{1}{2}\Delta dis_{i,g} - f_t(t) & \text{otherwise}
    \end{cases}
\end{equation}
\noindent where $f_t(\cdot) = \kappa_t \cdot t $ represents the penalty for algorithm time efficiency with respect to the penalty factor $\kappa_t \in [0, 1]$.

\section{Experiments And Results}
\subsection{Simulation Environment}

To address the MR-TASN task, we designed several simulated scenarios where a group of robots is assigned to search for individual target POIs in a human-filled environment, as shown in Fig.~\ref{fig:F1}. Formally, the MR-TASN task is modeled as a Dec-POSMDP, which couples the execution of temporally-extended macro actions for task allocation with the generation of primitive actions (LAs) for social navigation. The objective $\mathcal{J}$ and the blended MA\&LA policies are optimized using a MARL method, using a hypergraph-based network. 

In our MR-TASN simulator, robots are driven by a decentralized policy network trained using the CTDE paradigm. Pedestrians follows the state-of-the-art crowded behavior representation algorithm ORCA \cite{orca}, with agnostic personal strategies, where human velocity and goal positions are randomly changed. A set of static POIs is distributed between robots and humans, and robots are adaptively assigned to these POIs through MAs at each  decision-making timestep $t_k$. The initial states of robots, humans, and POIs are defined by distribution $\mathcal{S}_{0}$. The objective $\mathcal{J}$ is to maximize the expected macro-action reward, enabling robots to effectively access assigned POIs in order in highly dynamic, human-filled spaces while adhering to social norms.

\subsubsection{Baselines and Ablation Study}
we evaluate the performance of our framework, Hyper-SAMARL, by comparing it against several baseline methods and conducting an ablation study to assess the contribution of each component. We design a random-based averaged task allocation (RTA) method, which allocates tasks using a fully random function, to evaluate the allocation performance of Hyper-SAMALR. 

Two baseline MR-TASN algorithms are introduced: RTA-ORCA and RTA-A*. These are decoupled hierarchical methods with an RTA-based task allocator and either an ORCA-based \cite{orca} or A*-based \cite{A*} collision avoidance planner. 

Additionally, we include two ablation models in our experiments: RTA-SAMARL and MLP-SAMARL. RTA-SAMARL replaced the MA actor-critic network in Hyper-SAMARL with the RTA task allocation strategy, while keeping the same training parameters for socially-aware navigation planner, specifically the LA actor-critic network, as used in Hyper-SAMARL. MLP-SAMARL removes the Hypergraph neural network in Hyper-SAMARL and employs a multi-layer perceptron (MLP)-based actor-critic network for both task allocation and path planning.

\subsubsection{Training Details}
All the aforementioned algorithms are trained and tested under the same environmental configurations. Hyper-SAMARL, RTA-SAMRL and MLP-SAMARL are trained over a total of $1\times10^7$ timesteps, with a learning rate of $5\times10^{-4}$ for both the actor and critic networks. Other key parameters include PPO-epoch: 5, gain: 0.01, clipping factor: 0.2, and entropy coefficient: 0.01. 

\subsubsection{Evaluation}
\begin{figure}[!t]
\centering
\includegraphics[width=0.80\columnwidth]{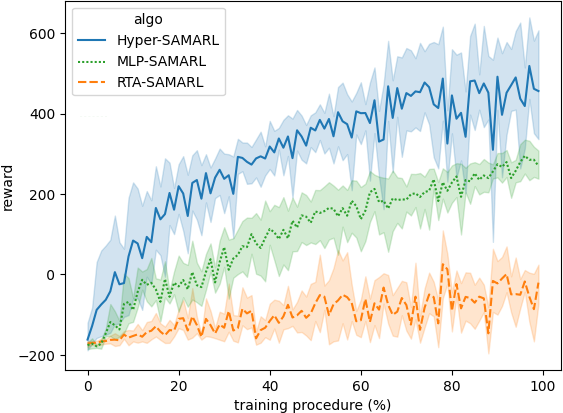}
\vspace{-4pt}
\caption{Learning curves of Hyper-SAMARL and other two ablation models with five different seeds.}
\vspace{-1pt}
\label{fig:curve}
\end{figure}

We have conducted our experiments on three different configurations: 1) 3 robots, 5 humans, 10 POIs; 2) 5 robots, 5 humans, 10 POIs; and 3) 5 robots, 10 humans, 20 POIs. Each algorithm was tested on a total of 500 random scenarios, where robots explore POIs that are randomly generated along a $50m \times 50m$ circle surrounding the human-generation circle in open space. We illustrate the learning curves of Hyper-SAMRL and the two ablation models in Fig.~\ref{fig:curve}, and compare two key metrics: average allocation score and average social score \cite{wang2024multi}, shown in Table~\ref{tab:table1}, to evaluate the overall performance of MR-TASN. 

\begin{table}[h]
\vspace{-5pt}
\caption{Simulation Experiment Results.\label{tab:table1}}
\vspace{-5pt}
\centering
\begin{scriptsize}
\begin{tabular}{cccccccc}
\hline & \multicolumn{3}{c}{ Average Allocation Score } & & \multicolumn{3}{c}{Average Social Score } \\
\cline { 2 - 4 } \cline { 6 - 8 } Methods & \multicolumn{3}{c}{ POI\&Robot\&Human } & & \multicolumn{3}{c}{ POI\&Robot\&Human } \\
& 10 & 10 & 20 & & 10 & 10 & 20 \\
&3\&5 & 5\&5 & 5\&10  && 3\&5 & 5\&5 & 5\&10 \\
\hline RTA-A* & $7$ & $14$ & $15$ & & $12$ & $11$ & $8$ \\
RTA-ORCA & $35$ & $40$ & $37$ & & $23$ & $20$ & $15$ \\
RTA-SAMARL & $27$ & $32$ & $23$ & & $82$ & $75$ & $70$ \\
MLP-SAMARL  & $62$ & $67$ & $59$ & & $54$ & $49$ & $41$\\
Hyper-SAMARL   & $\mathbf{93}$ & $\mathbf{95}$ & $\mathbf{91}$ & & $\mathbf{91}$ & $\mathbf{87}$ & $\mathbf{82}$\\
\hline
\end{tabular}
\end{scriptsize}
\vspace{-7pt}
\label{table:result}
\end{table}

The average allocation score estimates task allocation and is defined as: $\hat{\mathrm F}_{\mathrm{AS}} = 100 \times \frac{\sum_{i}^{n} n_{p}^{i}}{N_{P}}$, where $n_{p}^{i}$ is the number of POIs explored by the $i$-th robot, and ${N_{P}}$ denotes the total number of POIs. The average social score is adopted from the multi-robot social score \cite{wang2024multi}, who evaluates both the time efficiency of the path and the overall socially acceptable performance, considering the frequency and severity of uncomfortable interactions between pedestrians and robots. 

\subsubsection{Results}

As shown in Table~\ref{table:result}, the baseline RTA-A* exhibits limited performance in both the average allocation score (AS) and average social score (SC) metrics. This is because conventional static path planning methods, like RTA-A*, struggle to adapt effectively to dynamic environments. Another baseline, RTA-ORCA, performs better than the static collision avoidance planner RTA-A*, demonstrating a basic ability to handle dynamic environments. However, due to the short-sighted, one-step lookahead nature of traditional social navigation approaches, RTA-ORCA achieves a lower socially-acceptable score compared to learning-based approaches, especially under the pedestrian-invisible assumption. Although both RTA-ORCA and RTA-SAMARL use the same task allocation strategy, RTA-ORCA scores higher in AS. This is likely because learning-based planners like RTA-SAMARL typically take longer paths than ORCA, resulting in greater social acceptability but lower efficiency in task allocation.

For the ablation studies, RTA-SAMARL struggled with the task assignment (AS) metric due to its stochastic task allocation strategy for the robot group. An inefficient task allocation algorithm forces robots to waste time on unnecessary re-entrant routes. Moreover, although MLP-SAMARL shows balanced performance on the MR-TASN task, its limited HRI inference ability restricts further performance improvements compared to Hyper-SAMARL.

Notably, our proposed framework, Hyper-SAMARL, outperforms other baselines and ablation models in both the AS and SS metrics, as well as in reward collection, as shown in Table~\ref{table:result} and Fig.~\ref{fig:curve}. The use of a hypergraph-based neural network not only facilities the inference and understanding of complex environmental dynamics and potential correlations between objects but also enhances the adaptability of task assignments across different environmental conditions. Additionally, the cooperative strategies of the multi-robot system are integrated into the hypergraph-based neural network, developed through the MARL CTDE training procedure. In summary, the experimental results demonstrate that Hyper-SAMARL provides a benchmark-level performance for addressing the MR-TASN task in dynamic environments. Videos of the experiments on the MR-TASN task can be found \url{https://sites.google.com/view/hyper-samarl}.

\vspace{-3pt}
\section{Conclusion}
\vspace{-3pt}
We present Hyper-SAMARL, a framework for adjustable task allocation and coordinated socially-aware navigation with multi-robots using MARL and hypergraph neural networks. Hyper-SAMARL leverages a hypergraph neural network to optimize both a dynamic adaptable task allocation MA strategy and a social navigation LA planner, trained by MAPPO. Our results from simulations tests affirm its effectiveness, advancing multi-robot navigation. 




\typeout{}
\bibliography{main}
\bibliographystyle{IEEEtran}
\end{document}